# Statistical Translation, Heat Kernels and Expected Distances

**Joshua Dillon**
School of Electrical &
Computer Engineering
Purdue University
jvdillon@purdue.edu

**Yi Mao**
School of Electrical &
Computer Engineering
Purdue University
ymao@purdue.edu

**Guy Lebanon**
Statistics Department
Purdue University
lebanon@purdue.edu

**Jian Zhang**
Statistics Department
Purdue University
jianzhan@purdue.edu

## Abstract

High dimensional structured data such as text and images is often poorly understood and misrepresented in statistical modeling. The standard histogram representation suffers from high variance and performs poorly in general. We explore novel connections between statistical translation, heat kernels on manifolds and graphs, and expected distances. These connections provide a new framework for unsupervised metric learning for text documents. Experiments indicate that the resulting distances are generally superior to their more standard counterparts.

## 1 Introduction

Modeling text documents is an essential part in a wide variety of applications such as classification, clustering, segmentation, visualization and retrieval of text. Most approaches start by representing a document using the word histogram or term frequency (tf) representation. They then proceed to fit statistical models or compute distances based on the word histogram content. Representing a document by its word histogram may be motivated by the embedding principle, described in [19, 18]. It assumes that each document $x$ is generated by a certain multinomial $x \sim \text{Mult}(\theta_x)$ and that the multinomial parameter $\theta_x$ suffices to describe the data. In other words, we can turn our attention to quantities defined on the parameters, for example measure distances between documents $d(x, y)$ as distances between the corresponding parameters $d(\theta_x, \theta_y)$.

The framework described above embeds documents in the multinomial simplex

$$\overline{\mathbb{P}}_m = \left\{\theta \in \mathbb{R}^{m+1} : \forall i \ \theta_i \geq 0, \sum \theta_j = 1\right\}$$

which is the space of all multinomial parameters (we assume a vocabulary $V = \{1, \ldots, m+1\}$). We can thus define distances between documents as distances between the corresponding multinomial parameters $\theta, \eta \in \overline{\mathbb{P}}_m$, for example, the Euclidean distance $d(\theta, \eta) = \|\theta - \eta\|_2$ or the Fisher geodesic distance

$$d(\theta, \eta) = \arccos\left(\sum_{i=1}^{m+1} \sqrt{\theta_i \eta_i}\right) \qquad \theta, \eta \in \overline{\mathbb{P}}_m \quad (1)$$

which is based on the Fisher information metric [1].

The direct use of the multinomial parameters $\theta_x$ is problematic since these parameters are unobserved. We can use, however, estimated quantities for example the maximum likelihood estimator (mle) $\hat{\theta}_{\text{mle}}(x)$ which coincides with the word histogram or tf representation $[\hat{\theta}_{\text{mle}}(x)]_j \propto \sum_i \delta_{x_i,j}$. This motivation for the tf representation needs to be carefully examined due to the inefficient convergence of high dimensional histograms to their expectations. In other words, the word histogram of a document containing $N$ words serves as a poor estimate for the multinomial parameter $[\theta_x]_1, \ldots, [\theta_x]_{m+1}$ in the frequent case of $N \ll m$. Indeed, while $\hat{\theta}_{\text{mle}}(x)$ is unbiased, the total variance of its components grows linearly with $m$ resulting in high MSE error $\mathsf{E} \|\theta_x - \hat{\theta}_{\text{mle}}(x)\|^2$.

In this paper we propose a novel randomized representation for text documents that addresses the above problem. The representation uses word-to-word statistical translation to obtain randomized estimators of $\theta_x$ which are more accurate than the standard tf representation $\hat{\theta}_{\text{mle}}(x)$. This representation forms a random embedding in the multinomial simplex which leads to the concept of a random geometry. We demonstrate the usefulness of the concepts of expected distances and kernels in text classification.

The next section describes related work followed by a description of the translation model and random geometries. We conclude with a large deviation interpretation, experiments and a discussion.

## 2 Related Work

Our work is closely related to distributional clustering [22] which was first introduced to cluster words (such as nouns) according to their distributions in syntactic contexts (such as verbs). The model is estimated by minimizing various objective functions, and is used to address the data sparseness problem in language modeling. It also serves as an aggressive feature selection method in [2] to improve document classification accuracy. Unlike previous work, we use word contextual information for constructing a word translation model rather than clustering words.

Our method of using word translation to compute document similarity is also closely related to query expansion in information retrieval. Early work [24] used word clusters from a word similarity matrix for query expansion. A random walk model on a bipartite graph of query words and documents was introduced in [17], and was later generalized to a more flexible family of random walk models [10]. Noisy channels were originally used in communication for data transmission, but served as a platform for considerable research in statistical machine translation. An interesting work by Berger and Lafferty [4] formulated a probabilistic approach to information retrieval based upon the ideas and methods of statistical machine translation.

The idea of diffusion or heat kernel $\exp(-t\mathcal{L})$ based on the normalized graph Laplacian $\mathcal{L}$ [9] has been studied for discrete input space such as graphs [15], and applied to classification problems with kernel-based learning methods and semi-supervised learning [26, 3]. It has also been formally connected to regularization operators on graphs [23], and can be thought of as a smoothing operator over graphs. In our case the diffusion kernel is used to generate a stochastic matrix which is then used to define a translation model between words. This has the effect of translating one word to its semantic neighbor connected by similar contextual information.

Recently many topic models, such as Latent Dirichlet Allocation (LDA) [5], have been proposed to model text documents. Although most of them still assume a multinomial distribution for each document, the effective number of parameters could be much smaller. For example, in LDA the multinomial parameter for each document is actually a convex combination of the multinomial parameters of topic models. Thus the total number of parameters depends on the number of topics rather than number of documents.

Several methods have been proposed to learn a better metric for classification. In [25] a new metric is learned using side-information such as which pairs of examples are similar and dissimilar, and the task is formulated as a convex optimization problem. In [14] a quadratic Gaussian metric is learned based on the idea that examples in the same class should be collapsed, and the solution can be found again by convex optimization techniques. In [20] a new metric is estimated in an unsupervised manner based on the geometric notion of volume elements. A similar line of research develops new similarity measures and word clustering [6, 13] based on word co-occurrence information. In most methods, a linear transformation of the original Euclidean metric is learned using labeled data with criteria such as better separability or prediction accuracy. Unlike those methods, our approach is based on word translation and expected distances and is totally unsupervised.

## 3 Translation Model

Given a word in the vocabulary $v \in V = \{1, \ldots, m + 1\}$, we define its contextual distribution $q_v \in \overline{\mathbb{P}}_m$ to be $q_v(w) = p(w \in d | v \in d)$, where $p$ is a generative model for document $d$. In other words, assuming that $v$ occurs in the document, $q_v(w)$ is the probability that $w$ also occurs in the document. Note that in particular, $q_v(v)$ measures the probability of a word $v$ re-appearing a second time after its first appearance.

In general, we do not know the precise generative model and use standard estimates such as

$$\begin{aligned} q_w(u) &= \sum_{d'} p(u, d'|w) \\ &= \sum_{d'} p(u|d', w) p(d'|w) \\ &= \sum_{d'} \text{tf}(u, d') \frac{\text{tf}(w, d')}{\sum_{d''} \text{tf}(w, d'')} \\ &= \left( \frac{1}{\sum_{d'} \text{tf}(w, d')} \right) \left( \sum_{d'} \text{tf}(u, d') \text{tf}(w, d') \right) \end{aligned}$$

where $\text{tf}(w, d)$ is the relative (or normalized) frequency of occurrences of word $w$ in document $d$. Note that the above estimate requires only unlabeled data and can leverage large archival databases to produce accurate estimates.

As pointed out by several researchers, the contextual distributions $q_w, q_v$ convey important information concerning the words $w, v$. For example, similar distributions indicate a semantic similarity between the words. In this paper, we explore the geometric structure of the set of points $\{q_w : w \in V\} \subset \overline{\mathbb{P}}_m$ in order to define a statistical translation model that produces a better estimate of the document multinomial parameter.

Central to the translation model is the idea that re-placing occurring words with similar non-occurring words is likely to improve the estimate of the document's multinomial parameters while minimizing vocabulary mismatch. For example, a document about police activity should have large multinomial parameters, corresponding to the words `police` and `cops`, i.e., $[\theta_x]_{\text{police}} \approx [\theta_x]_{\text{cops}}$ and are large relative to the remaining parameters of $\theta_x$. The actual document $x \sim \text{Mult}(\theta_x)$ consists of a relatively small number of words and may only contain `police` and not `cops`. In some sense, the document should have contained `cops` as well but did not due to the fact that only a small number of words were sampled in the process of creating $x \sim \text{Mult}(\theta_x)$.

As a result of the above observation we stochastically translate a document $x$ into a new document $y$, where the word-to-word translation probabilities depend on the similarity between the contextual distributions. The multinomial $\theta_x$ is then represented by the word histogram of $y$ rather than of $x$. However, since the translation of the documents $x \to y$ is probabilistic, the estimator $\hat{\theta}_{\text{mle}}(y)$ is a random variable which in turn leads to derived random variables such as distances, geometries, and kernels.

### 3.1 Diffusion Kernel on $\{q_w : w \in V\}$

As mentioned above, the document translation is based on a word by word independent translation model where the translation probabilities $p(u \to w)$ are determined by some measure of the semantic similarity of the two words. Correlation between different vocabulary words is an inappropriate choice in our context since we are replacing words by other words rather than adding words to an existing document (as is done for example in query expansion). Similarity between contextual distributions $q_u, q_v$ indicates that $u$ and $v$ appear in the same context justifying the probabilistic replacement of $u$ with $v$.

The contextual distributions form a discrete set of points or a graph $G = (V, E)$, $V = \{q_w : w \in V\}, E \subset V \times V$ embedded in the manifold $\overline{\mathbb{P}}_m$. A natural probability function $p(u \to w)$ corresponding to probabilistic flow on the graph $G$ is given by the heat or diffusion kernel on the graph. Such a probability kernel is generally superior to $k$-step Markov transition probabilities since it aggregates probabilities of transitions along paths of varying lengths [9]. In fact, since the graph is embedded in the simplex $\overline{\mathbb{P}}_m$, the heat kernel may also be derived as the Riemannian heat kernel [16] restricted to a discrete set of points. In the above mentioned graphs we take as edge weights the heat

| jan | databas | nbc | wang | ottawa |
|-----|---------|-----|------|--------|
| feb | intranet | abc | chen | quebec |
| nov | server | cnn | liu | montreal |
| dec | softwar | hollywood | beij | toronto |
| oct | internet | tv | wu | ontario |
| aug | netscap | viewer | china | vancouv |
| apr | onlin | movi | chines | canada |
| mar | web | audienc | peng | canadian |
| sep | browser | fox | hui | calgari |

Figure 1: Lists of the 8 most similar words to each of the words in the top row. Similarity is computed based on the contextual distribution heat flow defined in Equation (2). Word endings are removed or slightly altered due the stemming process.

flow on the Fisher geometry of $\mathbb{P}_m$ [16]

$$e(u,v) = \exp\left(-\frac{1}{\sigma^2}\arccos^2\left(\sum_w \sqrt{q_u(w)q_v(w)}\right)\right). \quad (2)$$

The use of (2) as edge weights is motivated by the axiomatic derivation of the Fisher geometry [8, 7] and the experimental studies in [19]. Furthermore, Section 5 describes an interesting interpretation of the choice (2) through the use of the Chernoff-Stein Lemma.

The graph heat kernel is defined via the matrix exponential

$$T = \exp(-t\mathcal{L}) = I - t\mathcal{L} + \frac{1}{2!}t^2\mathcal{L}^2 - \frac{1}{3!}t^3\mathcal{L}^3 \cdots \quad (3)$$

of the normalized graph Laplacian [9]

$$\mathcal{L} = D^{-1/2}(D-E)D^{-1/2}.$$

Above, $E$ is the edge weight matrix of the graph and $D$ is a diagonal matrix with $D_{ii} = \sum_j e_{ij}$. The parameter $t$ corresponds to the time of heat flow on the graph or the amount of translation. Small $t$ would yield $T \approx I$ which implies no translation, while large $t$ would yield an approximately uniform $T$ which leads to translation between arbitrary words. Practical computation of the heat kernel $T$ is usually performed by computing the eigen-decomposition of $\mathcal{L}$, rather than by Equation (3).

Figure 1 displays lists of the the 8 most similar words to the five words appearing in the top row in a portion of the RCV1 corpus. Similarity is computed based on the contextual distribution heat flow defined in Equation (2). The corresponding translation model $p(u \to w)$ would, for example, translate semantically similar words to each other for example months of the year or Canadian cities.

## 4 Random Geometries and Expected Distances

The word-to-word translation $T_{u,w} = p(u \to w)$ defined in the previous section through the graph heat kernel gives rise to a random translation process $T^*$ on documents. Thus, instead of having an embedding $\hat{\theta}_{\text{mle}} : X \to \overline{\mathbb{P}}_m$ we have a probabilistic embedding $\hat{\theta}^{T^*}_{\text{mle}} = \hat{\theta}_{\text{mle}} \circ T^*$ and we obtain different points on the simplex with different probabilities. As a result, functions of the random embeddings such as distances

$$Y = d(\hat{\theta}^{T^*}_{\text{mle}}(x), \hat{\theta}^{T^*}_{\text{mle}}(y)) \qquad (4)$$

or kernels become random variables rather than scalars.

Random variables such as $Y$ in (4) can be used for modeling purposes in a number of ways. They induce a distribution which may be used to define a posterior distribution over the modeling task, for example predicted labels in classification. In cases where the posterior does not have a convenient analytic expression, sampling can be used to approximate the posterior mean. Alternatively, a more frequentist approach can be used in which a particular point estimate such as $\mathsf{E}(Y)$ is used, together with a confidence interval based on $\mathsf{Var}(Y)$, in a deterministic model fitting.

We concentrate below on computing the expected Euclidean distance for random embeddings which has a convenient closed form expression. We denote the histogram of a document $y = \langle y_1, \ldots, y_{N_y} \rangle$ as $\gamma(y)$ where $[\gamma(y)]_k = N_y^{-1} \sum_{i=1}^{N_y} \delta_{k,y_i}$. Using this notation, we have the following proposition.

**Proposition 1.** *Under the translation model described above,*

$$\mathsf{E}_{p(y|x)p(z|w)} \|\gamma(y) - \gamma(z)\|^2 = \qquad (5)$$

$$N_x^{-2} \sum_{i=1}^{N_x} \sum_{j \in \{1,\ldots,N_x\} \setminus \{i\}} (TT^\top)_{x_i, x_j} + N_x^{-1}$$

$$+ N_w^{-2} \sum_{i=1}^{N_w} \sum_{j \in \{1,\ldots,N_w\} \setminus \{i\}} (TT^\top)_{w_i, w_j} + N_w^{-1}$$

$$- 2 N_x^{-1} N_w^{-1} \sum_{i=1}^{N_x} \sum_{j=1}^{N_w} (TT^\top)_{x_i, w_j}$$

*where $T$ represents the heat kernel-based stochastic word-to-word translation matrix.*

*Proof.* First notice that

$$\mathsf{E}_{p(y|x)p(z|w)} \|\gamma(y) - \gamma(z))\|_2^2 = \mathsf{E}_{p(y|x)} \langle \gamma(y), \gamma(y) \rangle \qquad (6)$$
$$+ \mathsf{E}_{p(z|w)} \langle \gamma(z), \gamma(z) \rangle - 2 \mathsf{E}_{p(y|x)p(z|w)} \langle \gamma(y), \gamma(z) \rangle.$$

The proposition is proven by substituting in (6) the following expectations

$$\mathsf{E}_{p(y|x)p(z|w)} \langle \gamma(y), \gamma(z) \rangle$$
$$= N_x^{-1} N_w^{-1} \sum_{i=1}^{N_x} \sum_{j=1}^{N_w} \sum_{k=1}^{m+1} E_{p(y|x)p(z|w)} \delta_{k,y_i} \delta_{k,z_j}$$
$$= N_x^{-1} N_w^{-1} \sum_{i=1}^{N_x} \sum_{j=1}^{N_w} \sum_{k=1}^{m+1} T_{x_i, k} T_{w_j, k}$$
$$= N_x^{-1} N_w^{-1} \sum_{i=1}^{N_x} \sum_{j=1}^{N_w} (TT^\top)_{x_i, w_j}$$

$$\mathsf{E}_{p(y|x)} \langle \gamma(y), \gamma(y) \rangle$$
$$= N_x^{-2} \sum_{k=1}^{m+1} \sum_{i=1}^{N_x} \sum_{j=1}^{N_x} E_{p(y|x)} \delta_{k,y_i} \delta_{k,y_j}$$
$$= N_x^{-2} \sum_{i=1}^{N_x} \sum_{j \in \{1,\ldots,N_x\} \setminus \{i\}} (TT^\top)_{x_i, x_j}$$
$$+ N_x^{-2} \sum_{i=1}^{N_x} \sum_{k=1}^{m+1} T_{x_i, k}$$
$$= N_x^{-2} \sum_{i=1}^{N_x} \sum_{j \in \{1,\ldots,N_x\} \setminus \{i\}} (TT^\top)_{x_i, x_j} + N_x^{-1}$$

$\square$

It is worth mentioning several facts concerning the above expression. If $T = I$ the above expected distance reduces to the standard Euclidean distance between the histogram representations. While Equation (5) is expressed using sequential document contents, the expected distance remains the same under permutation of the words within a document since it is a pure bag of words construct. Finally, it is possible to pre-compute $TT^\top$ in order to speed up the distance computation.

Expression (5) can be used to compute the expected value of kernels, for example the linear kernel $\mathsf{E}_{p(y|x)p(z|w)} \langle \gamma(y), \gamma(z) \rangle$ and an RBF kernel. The resulting expectations can be shown to be symmetric positive definite kernels and can be used in any kernel machine.

## 5 A Large Deviation Interpretations

The method of types and the Chernoff-Stein Lemma provide an interesting interpretation to the heat kernel translation process. We present below a brief description. For more details refer to [12, 11].

**Proposition 2** (Chernoff-Stein Lemma). *Consider the hypothesis test for $X_1, \ldots, X_n \sim_{iid} Q$ between two alternatives $Q = P_1$ and $Q = P_2$ where $D(P_1 || P_2) < \infty$. Denoting the acceptance region for hypothesis $P_1$ by $A_n$ and the probabilities of error by $\alpha_n = P_1(A_n^c)$ and $\beta_n = P_2(A_n)$ we have*

$$\lim_{n \to \infty} \frac{1}{n} \log \min_{\alpha_n < \epsilon, A_n} \beta_n = -D(P_1 || P_2).$$

The above proposition indicates that the KL divergence $D(P_1 || P_2)$ is the best exponent in the probability of type II error while bounding the probability of type I error. Such a minimization of type II error while constraining type I error is standard practice in hypothesis testing. We thus have that the the optimal type II error behaves as $\beta_n^{\text{opt}} \approx \exp(-\gamma n D(P_1 || P_2))$.

The KL divergence $D(p||q)$ behaves at nearby points like the square of the Fisher geodesic distance $d^2(p,q)$. This can be easily seen by examining the second order Taylor series expansion of $D(p||q)$

$$D(p||q) = D(p||p) + \sum_x \frac{\partial D(p, p + \Delta)}{\partial \Delta(x)}\Big|_{\epsilon=0} \Delta(x)$$
$$+ \frac{1}{2} \sum_x \sum_y \frac{\partial^2 D(p, p + \Delta)}{\partial \Delta(x) \partial \Delta(y)}\Big|_{\epsilon=0} \Delta(x)\Delta(y) + o(\|\Delta\|^2).$$

Above, the first order terms zero out and the second order terms become the squared length element of $\Delta$ under the Fisher metric justifying the relationship $2D(p||q) \approx d^2(p,q)$ for nearby $p,q$. More precisely, it can be shown that

$$\lim_{q \to p} \frac{d^2(p,q)}{2D(p||q)} = 1$$

with uniform convergence as $d(p,q) \to 0$.

Combining this result with the Chernoff-Stein Lemma we get a striking interpretation of our heat kernel translation model. The heat kernel is based on a graph whose edge weights (2) approximate the optimal error rate of the hypothesis test between the alternatives $Q = q_u$ and $Q = q_v$. In other words, the probability of translating $u \to v$ is directly specified by the optimal error rate or difficulty in distinguishing between the two contextual distributions $q_u, q_v$.

## 6 Experimental Results

In this section, we demonstrate the effect of using the expected $L_2$ distance vs. the $L_2$ distance in the context of nearest neighbor text classification and kernel PCA dimensionality reduction using the Reuters RCV1 corpus.

### 6.1 Nearest Neighbor Classification

For nearest neighbor classification, various 1 vs all binary classification tasks were generated from the RCV1 topic hierarchy. Figure 2 lists all such tasks, which are the top 5 largest leaf categories from each C,E,G,M root category. The reduction in the error rate of replacing $L_2$ with its expected version are demonstrated for balanced training sets of sizes equal to 100, 200 and 500. The testing documents are also balanced and fixed to be 200. The averaged results from 40 realizations are positive in general and indicate an overall tendency for significant improvement.

The top 2000 words (words that appear in most documents) were excluded from participating in the translation. This exclusion follows from the motivation of translation as obtaining more accurate estimates of low frequency terms. The most common terms already appear often and their corresponding multinomial estimates are accurate. There is no need to translate from them to other words and vice versa. It is important to realize that this exclusion does not eliminate or downweight the frequent words like the tfidf representation in any way. It merely limits the translations between these words and other words. All parameters, such as heat kernel $t$, were chosen experimentally.

### 6.2 Embedding a Random Geometry through Kernel PCA

We also explore the random geometry introduced in section 4 in the context of dimensionality reduction. A commonly used approach for projecting data into a lower dimensional subspace is principal component analysis (PCA). This technique attempts to represent the data on a set of orthonormal axes in such a way that minimizes the global sum-square reconstruction error. As such, subsequent coordinates of this projection capture decreasing levels of empirical variance. Naively applying PCA to the random geometry, one must compute the expected covariance matrix between the words of all documents, then use the dominant eigen vectors of this matrix as the projection space. In practice however, we resort to the kernel PCA method because of its ability to capture nonlinear structure as well its reduced computational overhead. To this end, we have explored two expected kernels, both of which are natural extensions of the linear and RBF kernels to the random geometry. The expected linear kernel, defined as $K(\gamma(x), \gamma(y)) = E_{p(y|x)p(z|w)} \langle \gamma(y), \gamma(z) \rangle$, while the expected RBF kernel is obtained by replacing the squared $L_2$ distance with the expected squared $L_2$ distance: $K(\gamma(x), \gamma(y)) = \exp\left(-\frac{1}{\sigma^2} E_{p(y|x)p(z|w)} \|\gamma(y) - \gamma(z))\|_2^2\right)$. Both kernels are symmetric and positive definite and can be used

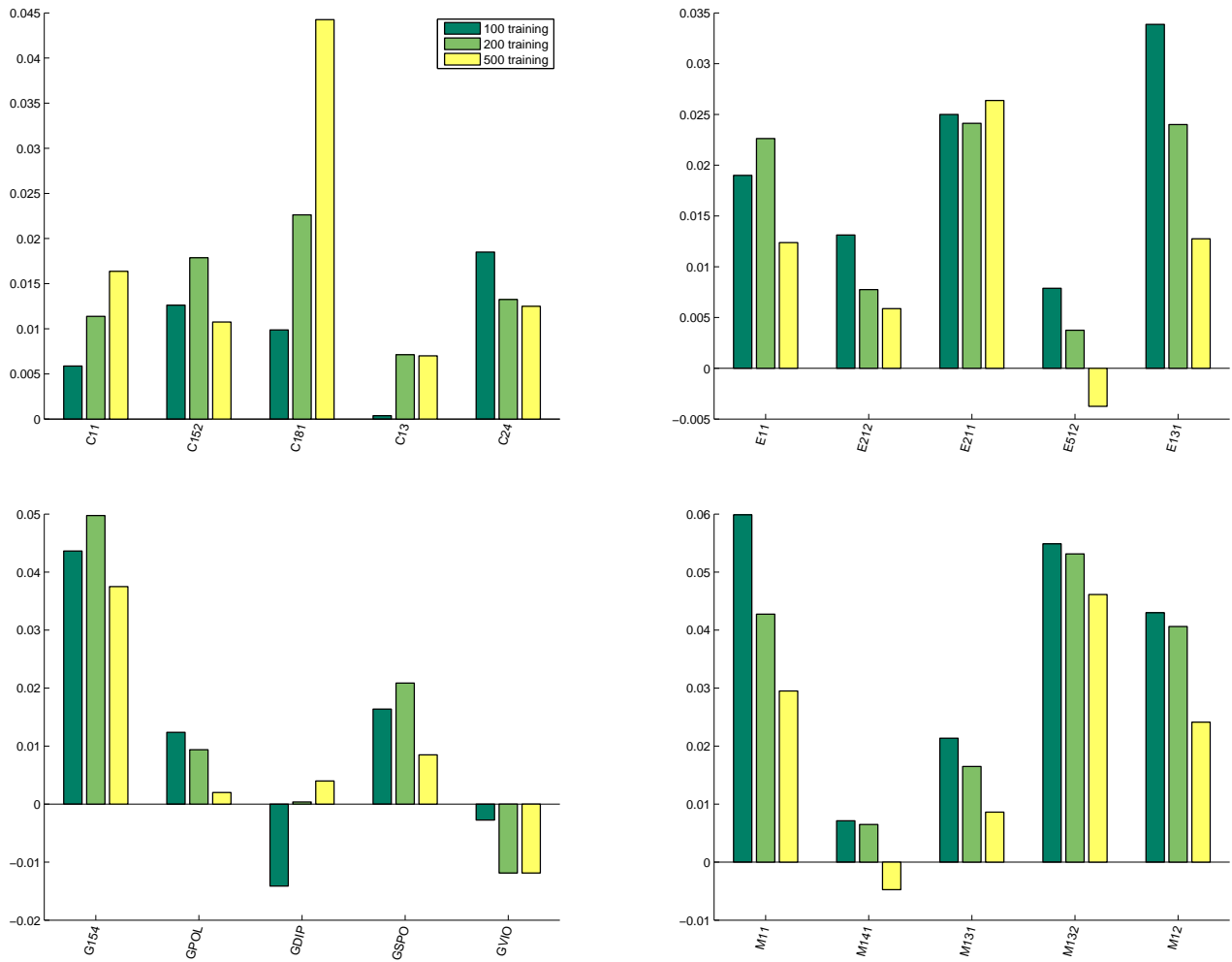

Figure 2: Average error reduction of expected $L_2$ from $L_2$ over 40 realizations of balanced training and testing documents for 20 RCV1 topic categories with a nearest neighbor classifier.

in any kernel machine.

Table 2 and 3 attempt to quantify the performance of the kernel PCA under expected linear and RBF kernels. To this end, we formed four different multiple label tasks, C, E, G, and M (Table 1). Each task contains a set of topics from RCV1 categories. During each iteration, we randomly pick 3000 uniquely labeled documents from the corpus, forming a 70/30 train-test split. We employ the following criteria to evaluate performance: (1) classification error rate of Fisher's linear discriminant when documents are first projected to the $k$-dimensional eigen-subspace; (2) the percentage of the overall variance captured by the first $k$ eigenvalues for training data; (3) the extent to which the projection captures new data (measured by the average sum-of-square residual of the test data). All the numbers are averaged from 25 cross validations. Good performance correspond to low numbers in the first

| task | topics from RCV1 |
|---|---|
| C | C152 C181 C13 C24 C21 |
| E | E212 E211 E512 E131 E11 |
| M | M11 M12 M131 M132 M141 M142 M143 |
| G | GPOL GDIP GSPO GVIO GCRIM |

Table 1: Topics from RCV1 that are involved in the kernel PCA experiments.

and third tasks and high numbers in the second task.

We report our results for eigen-subspaces of sizes $k = 1, 2, 5, 10$. The expected linear kernel always outperforms the linear kernel under current experimental setup. The expected RBF kernel performs better in most of the cases, but it shows an increase in the test data residual for task G and M. Also, for the tasks under consideration, the (expected) RBF kernels performs substantially worse than the (expected) linear kernels which may be a consequence of not trying enough possible values for the RBF scale parameter $\sigma^2$.

|   | $k=1$ | $k=2$ | $k=5$ | $k=10$ |
|---|---|---|---|---|
| | multiclass classification error rate | | | |
| C | 0.7443 (0.7540) | 0.5836 (0.6040) | 0.5099 (0.5252) | 0.3798 (0.3874) |
| E | 0.5872 (0.6271) | 0.4436 (0.4717) | 0.3557 (0.3976) | 0.1852 (0.1973) |
| G | 0.4601 (0.4912) | 0.3873 (0.4216) | 0.2718 (0.3080) | 0.1914 (0.2005) |
| M | 0.7075 (0.7327) | 0.5685 (0.6356) | 0.4236 (0.4356) | 0.3287 (0.3465) |
| | portion of the variance captured | | | |
| C | 0.1072 (0.0928) | 0.1424 (0.1234) | 0.2125 (0.1854) | 0.2797 (0.2459) |
| E | 0.0820 (0.0708) | 0.1494 (0.1302) | 0.2544 (0.2276) | 0.3376 (0.3066) |
| G | 0.0369 (0.0292) | 0.0576 (0.0462) | 0.1022 (0.0831) | 0.1531 (0.1256) |
| M | 0.0433 (0.0372) | 0.0740 (0.0638) | 0.1484 (0.1277) | 0.2335 (0.2009) |
| | average residual of the test data | | | |
| C | 0.0264 (0.0308) | 0.0250 (0.0294) | 0.0229 (0.0273) | 0.0211 (0.0254) |
| E | 0.0245 (0.0285) | 0.0227 (0.0266) | 0.0197 (0.0235) | 0.0176 (0.0211) |
| G | 0.0173 (0.0227) | 0.0170 (0.0224) | 0.0163 (0.0216) | 0.0156 (0.0209) |
| M | 0.0288 (0.0339) | 0.0281 (0.0332) | 0.0261 (0.0312) | 0.0238 (0.0288) |

Table 2: Results are averaged over 25 realizations of 2100 training and 900 testing documents for RCV1 C, E, G, M categories using kernel PCA with expected linear kernel (and regular PCA).

|   | $k=1$ | $k=2$ | $k=5$ | $k=10$ |
|---|---|---|---|---|
| | multiclass classification error rate | | | |
| C | 0.7448 (0.7551) | 0.7758 (0.8068) | 0.8381 (0.8446) | 0.7644 (0.7712) |
| E | 0.7218 (0.7673) | 0.8140 (0.8750) | 0.7259 (0.7466) | 0.5868 (0.6222) |
| G | 0.6990 (0.7039) | 0.6982 (0.7039) | 0.5846 (0.6506) | 0.6441 (0.6714) |
| M | 0.7487 (0.7569) | 0.7276 (0.7445) | 0.7070 (0.7361) | 0.5982 (0.6394) |
| | portion of the variance captured | | | |
| C | 0.1066 (0.0919) | 0.1411 (0.1222) | 0.2109 (0.1836) | 0.2777 (0.2431) |
| E | 0.0816 (0.0693) | 0.1486 (0.1287) | 0.2534 (0.2249) | 0.3366 (0.3033) |
| G | 0.0381 (0.0302) | 0.0584 (0.0468) | 0.1024 (0.0835) | 0.1533 (0.1255) |
| M | 0.0431 (0.0369) | 0.0740 (0.0634) | 0.1480 (0.1261) | 0.2318 (0.1986) |
| | average residual of the test data | | | |
| C | 0.9803 (0.9859) | 0.9535 (0.9645) | 0.8986 (0.9102) | 0.8353 (0.8537) |
| E | 0.9827 (0.9902) | 0.9529 (0.9591) | 0.9176 (0.9306) | 0.8828 (0.8920) |
| G | 0.8740 (0.8701) | 0.8687 (0.8672) | 0.7804 (0.7876) | 0.7061 (0.7094) |
| M | 0.9797 (0.9902) | 0.8306 (0.8141) | 0.7699 (0.7434) | 0.6577 (0.6054) |

Table 3: Results are averaged over 25 realizations of 2100 training and 900 testing documents for RCV1 C, E, G, M categories using kernel PCA with expected RBF kernel (and regular RBF kernel).

## 7 Discussion

The experimental results demonstrate how overall the ideas of random embeddings via statistical translation and heat kernels combine to contribute to more accurate estimation. The large deviation interpretation of the heat kernel translation draws another interesting connection and may lead to additional intriguing results.

It is likely that expected distances may be used in other areas of text analysis such as query expansion in information retrieval. Another interesting application that is worth exploring is using the expected distances in sequential visualization of documents, for example based on the lowbow framework [21].

The theoretic motivation behind the translation model as a probabilistic biased estimate of the document multinomial parameters should be further explored. A solid connection, if found, between translation based expected distances and variance reduction would be an interesting result to the statistics and machine learning communities. It would link theoretical results in statistical estimation in high dimensions to practical information retrieval methods such as query expansion.

In practice our model works best without translating frequent words (see explanation in Section 6.1). Future work is expected to provide a more rigorous study, for example by forming a prior on the translation of each word that depends on the confidence intervals of the corresponding multinomial parameter estimation.

In general, the standard practice of treating words as orthogonal for classification and visualization is ill suited for short documents and large vocabulary size. In this work, we use an unlabeled corpus to learn a semantic structure leading to a more effective parameter estimation and more robust comparison of documents.